\pdfoutput=1
\documentclass[11pt]{article}
 
\usepackage[table]{xcolor}
\usepackage{acl}
\usepackage{times}
\usepackage{latexsym}

\usepackage[T1]{fontenc}
\usepackage[utf8]{inputenc}

\usepackage{microtype}

\usepackage{inconsolata}

\usepackage[noend]{algpseudocode}
\usepackage{subcaption}
\usepackage{natbib}
\usepackage{multicol}
\usepackage{multirow}
\usepackage{array}
\usepackage{tabu}
\usepackage{booktabs}
\usepackage{enumitem}
\usepackage{graphics}
\usepackage{graphicx}
\usepackage{amssymb}
\usepackage{amsmath}
\usepackage{mathrsfs}
\usepackage{amsfonts}
\usepackage{algorithmicx}
\usepackage[ruled, lined, linesnumbered, commentsnumbered, longend]{algorithm2e}
\definecolor{applegreen}{rgb}{0.55, 0.71, 0.0}

\usepackage{amsmath,amsfonts,bm}

\def\eqref#1{equation~\ref{#1}}
\def\1{\bm{1}}

\DeclareMathAlphabet{\mathsfit}{\encodingdefault}{\sfdefault}{m}{sl}
\SetMathAlphabet{\mathsfit}{bold}{\encodingdefault}{\sfdefault}{bx}{n}

\def\gA{{\mathcal{A}}}

\def\gC{{\mathcal{C}}}

\def\gE{{\mathcal{E}}}

\def\gG{{\mathcal{G}}}

\def\gI{{\mathcal{I}}}

\def\gL{{\mathcal{L}}}

\def\gP{{\mathcal{P}}}

\def\gR{{\mathcal{R}}}

\def\gV{{\mathcal{V}}}

\def\sD{{\mathbb{D}}}
\def\sG{{\mathbb{G}}}

\def\sN{{\mathbb{N}}}

\def\sR{{\mathbb{R}}}
\def\sS{{\mathbb{S}}}

\def\sV{{\mathbb{V}}}
\def\sW{{\mathbb{W}}}

\newcommand\me{$\textsc{MEEL}$}
\newcommand\mev{$\textsc{M-EV}^{2}$}

\title{\me: Multi-Modal Event Evolution Learning}

\author{Zhengwei Tao$^{12}$ ~Zhi Jin$^{12}$\thanks{*Corresponding authors.} ~Junqiang Huang$^3$ ~Xiancai Chen$^{12}$  ~Xiaoying Bai$^{4*}$ \\{\bf ~Haiyan Zhao$^{12}$  ~Yifan Zhang$^{12}$ ~Chongyang Tao$^5$}\\
        $^1$Key Laboratory of High Confidence Software Technologies (PKU), MOE, China\\
        $^2$School of Computer Science, Peking University ~$^3$Peking University\\
        $^4$Advanced Institute of Big Data ~$^5$Beihang University\\ 
        \texttt{\{tttzw, xiancaich, yifanzhang\}@stu.pku.edu.cn}, ~\texttt{baixy@aibd.ac.cn}\\ ~\texttt{\{zhijin,zhhy.sei\}@pku.edu.cn}
        ~\texttt{chongyang@buaa.edu.cn}}

\begin{document}
\maketitle
\begin{abstract}

Multi-modal Event Reasoning (MMER) endeavors to endow machines with the ability to comprehend intricate event relations across diverse data modalities. 
MMER is fundamental and underlies a wide broad of applications. 
Despite extensive instruction fine-tuning, current multi-modal large language models still fall short in such ability. The disparity stems from that existing models are insufficient to capture underlying principles governing event evolution in various scenarios. 
In this paper, we introduce Multi-Modal Event Evolution Learning (\me) to enable the model to grasp the event evolution mechanism yielding advanced MMER ability. Specifically, we commence with the design of event diversification to gather seed events from a rich spectrum of scenarios. Subsequently, we employ ChatGPT to generate evolving graphs for these seed events. We propose an instruction encapsulation process that formulates the evolving graphs into instruction-tuning data, aligning the comprehension of event reasoning to humans. 
% To further improve the evolution process, we implement the guiding discrimination paradigm, in which we design various event negative mining strategies and train the model in precise event classification. 
Finally, we observe that models trained in this way are still struggling to fully comprehend event evolution. In such a case, we propose the guiding discrimination strategy, in which models are trained to discriminate the improper evolution direction.
We collect and curate a benchmark \mev~for MMER. Extensive experiments on \mev~validate the effectiveness of our approach, showcasing competitive performance in open-source multi-modal LLMs.
Code and Dataset are available on \href{https://github.com/TZWwww/MEEL}{https://github.com/TZWwww/MEEL}.

\end{abstract}
   
\section{Introduction}
\label{sec:intro}

% 一
% 1. event是什么
% 2. mm event reasoning是什么
% 3. mm event reasoning有什么任务
% 4. mm event reasoning有什么意义

% 二
% 1. 随着大语言模型研究的深入，在经过指令微调的纯文本模态的LLM已经能很好地处理纯文本事件推理。
% 2. MMER相较于纯文本事件推理是更复杂的任务。他要求模型能同时理解来自不同模态的事件信息，然后将其对齐，最后再展开综合推理。
% 3. 在我们的实验中，我们发现当前的多模态大语言模型尽管经过了大量的指令微调，但仍然不能完成复杂的MMER任务。
% 4. 其原因是多方面的。首先，事件的演化连续的，当前的多模态LLM的指令微调数据仅包含对事件演化局部的推理，难以使模型具备对事件演化整体结构的理解能力。其次，现有多模态LLM没有建模事件关系之间的相互作用，导致模型难以推理复杂的复合事件关系。

% 三 (写的很简单，后面再扩充)
% 1. 为了解决这个问题，我们提出Multi-Modal Event Evolution Learning.
% 2. 具体来说，为了解决上述的第一个问题，我们首先利用ChatGPT生成事件的演化图，用以增强模型的事件演化理解。
% 3. 针对第二个问题，基于这个演化图，我们设计了事件关系的组合归纳，训练模型能够推理复杂的复合事件关系。
% 4. 我们构建了面向多模态事件推理的指令微调数据集，训练了能更好完成多模态事件推理的LLM。

% 四
% 为了验证我们方法的有效性，我们在7个数据集上做了实验。实验结果表明我们的方法取得了开源多模态事件关系推理的SOTA。

%-------------------------------------------------------------------------

Events are instances or occurrences that are the fundamental semantic units.
% Multi-modal event reasoning (MMER) endeavors to empower machines with the capability to comprehend the interconnections of causality, temporality, and intent among events, leveraging both visual and textual information.
Events are not independent, and they are usually interconnected by the following relations: causality, temporality, and intention. Multi-modal Event Reasoning (MMER) is to comprehend these events and their relations in both visual and textual modalities, and finally pave a path to better understanding the true world.
MMER is expected to serve as the underpinning for various multi-modal applications, including visual storytelling~\cite {huang2016visual}, visual event prediction~\cite{huang2021igseg}, event-related VQA~\cite{park2020visualcomet}, MM knowledge graph construction~\cite{ma2022mmekg}, and video generation~\cite{li2018video,liu2024sora}.
Such intricate tasks require an understanding of the event evolution mechanism across diverse scenarios.

\begin{figure}[!tb]
\setlength{\belowcaptionskip}{-5mm}
    \centering
    \includegraphics[width=1\columnwidth]{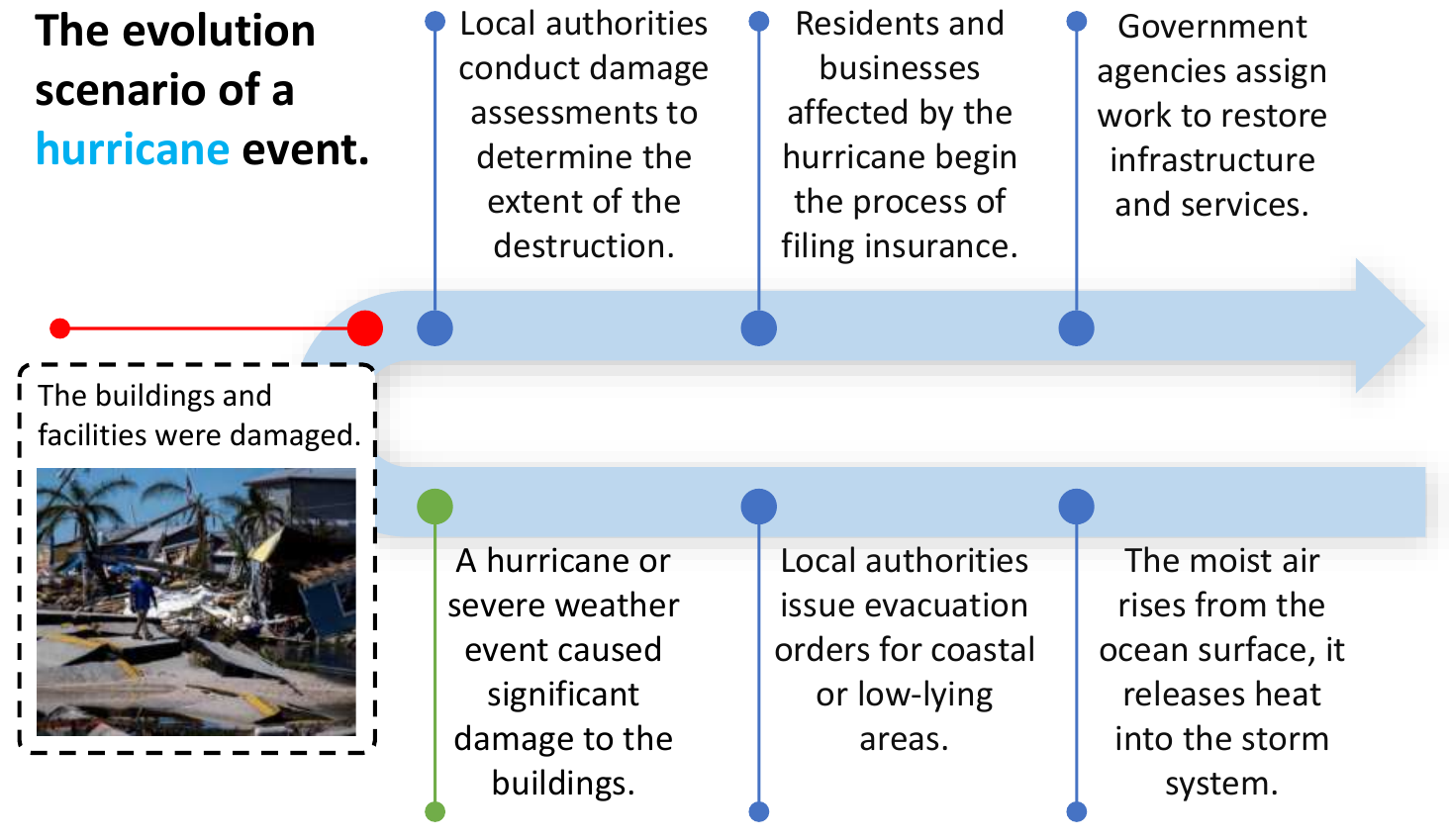}
    \caption{Part of the event evolution of a hurricane scenario. The queried event is in {\color{red}red}. \me~endows the model with the knowledge of all events in the scenario evolution. Current methods only train the model of few clips of event reasoning of the {\color{applegreen}green} one. }
    \label{fig:intro}
\end{figure}

With the deepening of research on multi-modal instruction tuning, Multi-modal large language models (MLLM) have been able to handle various multi-modal tasks effectively~\cite{liu2023visual, zhu2023minigpt, chen2023minigpt, dai2023instructblip, li2023otter}. 
These models master some abilities of MM event reasoning implicitly during training in diversified sorts of tasks. 
Among all the task categories, the perception tasks such as referring expression comprehension, referring expression generation, and grounded image captioning~\cite{mao2016generation,kazemzadeh2014referitgame,peng2023kosmos} enable the model to comprehend the entities of the events in the image and text. 
The cognitive tasks, namely image caption and VQA~\cite{lin2014microsoft, goyal2017making}, endow the model with the semantic understanding capability of events. 
However, the models trained by these tasks are unable to perceive event evolution because of the static nature of all modality inputs. 
% However, these event-evolution agnostic tasks are insufficient for models to graph the event-evolving mechanism.
Existing visual instruction-tuning methods only consist of questions for few clips of the entire event scenario. 
As shown in Figure~\ref{fig:intro}, current methods only model the queried events with the green event and ignore the rest of the scenario. 
They lack a vision of a broad spectrum of other events in the evolving context. 
Such contextual absence impedes models from learning abundant evolution knowledge resulting in poor performances in MMER.

% With the deepening of research on large language models, the instruction-tuned textual modality of the LLM has been able to handle pure-text event reasoning effectively.
% Compared to pure-text event reasoning, MMER is a more complex task. It requires the model to understand event information from different modalities simultaneously, align them, and then unfold comprehensive reasoning.
% In our experiments, we found that current multimodal large language models, despite extensive instruction-fined, still cannot complete complex MMER tasks.
% The reasons are manifold. 
% Firstly, The evolution and development of events have their own underlying principles, leading to events happening in graph structure~\cite{zhang2022aser}. Such structure can reveal the basic rules of event evolution of a real-world scenario.
% The current multi-modal LLM only trains to reason in shallow event evolution of limited scenarios. These models fail to understand the entire event scenario and lack rich knowledge of event evolution.
% Secondly, existing models are not trained for event evolution discrimination. These models are not able to be aware of their internalized false event evolution knowledge. As a result, they suffer from reasoning hallucinations~\cite{zhang2023siren}.

To address this issue, we propose Multi-Modal Event Evolution Learning~(\me) for endowing the model to understand the event evolution to enhance the ability of MMER, leading to improved performances on downstream tasks.
Specifically, we first design the event scenario diversification to acquire various events from abundant scenarios.
Then, we employ ChatGPT to generate the evolving graphs of these seed events.
The aim is to use these graphs to train the model to understand the rich knowledge of the evolution of events.
To accomplish this goal, we propose the instruction encapsulation process to adapt the evolving graphs into instruction-tuning data to train the model. 
In this way, the training allows the model to comprehend more event evolutional knowledge of the scenario leading to better performance of MMER. 
However, allowing the model to learn only the evolving graphs is insufficient. 
Without acknowledging the incorrect evolving events, the model would improperly forward the process, resulting in the hallucination of event reasoning.
To mitigate this problem, we perform the guiding discrimination. The model requires judging the incorrect evolution. We design various negative mining strategies to harvest incorrect events. Then, we train the model to classify the right event. 
We also adapt the guiding discrimination into instruction tuning.
After obtaining all the data, We finetune the LLaVA~\cite{liu2023visual} model after its stage-1 pre-taining with LoRA~\cite{hu2021lora} to get our model.

To validate the effectiveness of \me, we curate a benchmark \mev~for \textbf{M}ulti-modal \textbf{EV}aluation of \textbf{EV}ent reasoning. \mev~is collected or curated from nine existing datasets covering visual storytelling~\cite{huang2016visual}, visual event prediction~\cite{huang2021igseg}, and event-related VQA~\cite{yeo2018visual, zhang2021learning}. \mev~consists tasks relying on the abilities of MMER of diverse inter-event relations as causality, temporality, and intent. It also consists of two reasoning paradigms: close and open reasoning.
We conducted extensive experiments on \mev~and compare \me~to MLLM baselines.
We achieve competitive performances in open-source MM LLMs.
The results demonstrate that our method does enhance the MMER ability of the model yielding significant improvements in downstream tasks.
We conclude our contributions as:

\begin{figure*}[!tb]
    \centering
    \includegraphics[width=2\columnwidth]{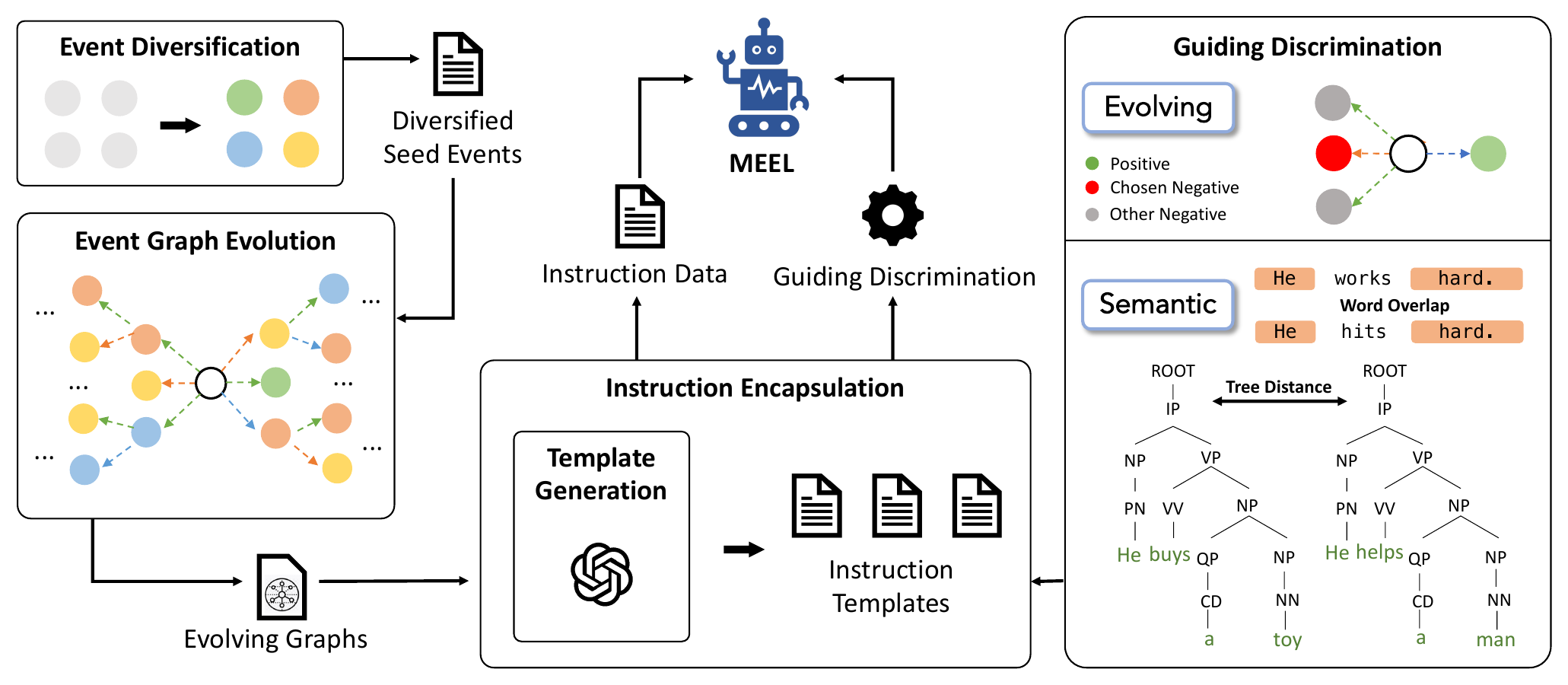}
    \caption{Overview of \me. We first implement the Event Diversification to harvest seed events. Then we perform the Event Graph Evolution to obtain the evolving graphs. We adapt the evolving graphs into instruction-tuning data through our Instruction Encapsulation. The Guiding Discrimination aims to improve the evolution learning with our two negative event mining strategies. }
    \label{fig:overview}
\end{figure*}

\begin{itemize}[topsep=0pt]
\setlength{\itemsep}{1pt}
\setlength{\parskip}{0pt}
\setlength{\parsep}{-1pt}
\setlength{\leftmargin}{-1pt}
\item[$\bullet$] We propose the Multi-Modal Event Evolution Learning~(\me). It aims to train the model to comprehend the intricate event evolution of diversified scenarios. Our method may shed light on other MM event reasoning research. 
    
\item[$\bullet$] We further design the Guiding Discrimination to guide the evolution and mitigate the hallucinations of MMER.
    
\item[$\bullet$] We collect and curate the \mev~benchmark for MMER. \mev~covers diversified inter-event relations. We conduct extensive experiments on \mev~to test the effectiveness of our model. We achieve competitive performance among open-source MLLMs.
\end{itemize}

\section{Multi-Modal Event Evolution Learning}

% Our purpose is to train a multi-modal large language model that endows with the advanced ability of multi-modal event reasoning~(MMER) leading to improved performances on downstream tasks. 
% We propose Multi-Modal Event Evolution Learning~(\me) to achieve our goal. We introduce the formulation of MMER (Section~\ref{mmer}). For our method \me, we first design the event diversification process to obtain the seed events with enriched types and scenarios~(Section \ref{ed}). We then perform the event graph evolution to harvest the event-evolving graphs of the seed events~(Section \ref{ege}). We aim to train the model with the knowledge of the event-evolving graphs. Therefore, we conduct instruction encapsulation to adapt the event-evolving graphs into instruction-tuning data~(Section \ref{ie}). In order to guide for better evolution and mitigate the reasoning hallucinations, we propose the guiding discrimination training paradigm~(Section \ref{eld}). We show the overview of \me~in Figure~\ref{fig:overview}.

% ===
We strive to enhance a multi-modal large language model's capability in multi-modal event reasoning (MMER) to boost performance on downstream tasks. Our approach, Multi-Modal Event Evolution Learning (\me), is introduced and structured as follows: Section~\ref{mmer} details the MMER task. The main purpose of \me~is to enhance the comprehension of event evolution.
We initiate with an event diversification step to generate a diverse mix of seed events of various scenarios (Section \ref{ed}). Then we construct the event-evolving graphs through a novel method named event graph evolution (Section \ref{ege}). Our next objective is to leverage these event-evolving graphs for model training. To this end, we encapsulate these graphs into suitable formats for instruction tuning by instruction encapsulation (Section \ref{ie}). Note that instruction tuning is one of the feasible ways to learn the knowledge of event-evolving graphs. One can also leverage other methods such as in-context learning. Finally, we incorporate a guiding discrimination training strategy to refine evolution pathways and reduce reasoning errors (Section \ref{eld}). \me's comprehensive framework is graphically represented in Figure~\ref{fig:overview}.

% \subsection{Multi-Modal Event Reasoning}
% \label{mmer}
% Events are defined as basic semantic molecules to explain the states or actions of peoples and entities, or things happening in a specific scenario. Event semantics can be better captured by event-associated images rather than merely by the text description of it~\cite{zhang2021merl}. Multi-Modal Event Reasoning~(MMER) aims to reason multi-modal events via various inter-event relations (e.g., temporal, causal, intentional), thus enabling commonsense or cognitive reasoning capabilities~\cite{tao2023unievent,tao2023eveval,han2021ester}. MMER underlies a wide sorts of downstream applications~\cite{huang2016visual, park2020visualcomet, huang2021igseg}.

% We introduce the formulation of MMER in detail. An event can be described by a sentence $\gE$ and depicted by an image $\gI$. On one hand, the sentence describes it with its arguments such as the subject, the verb, and the object~\cite{ace2004}. On the other, the image would depict the surrounding environment and situation information~\cite{yang2023event,zellers2021merlot}. 
% Given an interested relation $\gR$, MMER can be formulated as reasoning the target event $\gE^{t}$ according to $\gR$:
% \begin{equation}
% \label{gen}
%     \gE^{t} = \mathrm{M}~(\gE, \gI, \gR), \quad \gR\in\sS^{\gR}
% \end{equation}
% $\mathrm{M}$ represents the reasoning model. $\sS^{\gR}$ is the universe of the inter-event relations. $\gE$ or $\gI$ could be either absent.

% ===

\subsection{Multi-Modal Event Reasoning}
\label{mmer}
Multi-Modal Event Reasoning (MMER) involves deducing events based on certain inter-event relations across different modalities. 
% This is the fundamental abilities enhancing comprehension of events that are described by both textual and visual elements. 
Specifically, events as semantic units can be characterized by text, but their semantics are often more richly conveyed through associated images~\cite{zhang2021merl}. The pursuit of MMER is to harness these multi-modal inputs to establish various relationships between events (temporal, causal, intentional, etc.), facilitating sophisticated reasoning processes~\cite{tao2023unievent,tao2023eveval,han2021ester}. This reasoning underlies a spectrum of downstream tasks~\cite{huang2016visual, park2020visualcomet, huang2021igseg}.

We elaborate on the MMER formulation, wherein an event is expressed by a textual sentence $\gE$ and represented by an image $\gI$. Text provides argument structure, such as subject, verb, and object~\cite{ace2004}, while images contextualize the event with environmental and situational details~\cite{yang2023event,zellers2021merlot}. MMER can be modeled as inferring a target event $\gE^{t}$ based on a given relation $\gR$:
\begin{equation}
\label{gen}
    \gE^{t} = \mathrm{M}~(\gE, \gI, \gR), \quad \gR\in\sS^{\gR}.
\end{equation}
Here, $\mathrm{M}$ denotes the model and $\sS^{\gR}$ represents the set of possible inter-event relations. 
For example, in Figure~\ref{fig:intro}, $\gE$ is the red event, $\gI$ is the image, the queried relation $\gR$ is "cause", the answer $\gE^{t}$ is the green event. Therefore, the entire data is:

Question: \textit{Given the image, what is the cause of "The buildings and facilities were damaged.".}

Answer: \textit{A hurricane or severe weather
event caused significant damage to the buildings.}

This question can not be answered only based on the $\gE$ since there can be many reasons for building damage. Seeing the image, we can reason the damage could be caused by a hurricane.
Models require analysis of both $\gE$ and $\gI$ to get the answer.
% Sometimes, the task allows for the absence of either $\gE$ or $\gI$ in context.

% \subsection{Event Diversification}
% \label{ed}
% The event diversification is designed to obtain the events of assorted types and scenarios. Then we regard these events as the seed events for the following evolution.

% We begin with a set of text and image pairs $\{(\gE_i, \gI_i)\}$, $\gE_i$ and $\gI_i$ together represent the event. 
% Since the event’s trigger is the word that most clearly expresses its
% occurrence~\cite{ace2004}, we extract the trigger verb to reflect its type and scenario. For each $\gE_i$, we extract its first verb~$\gV_{\gE_i}$ by utilizing Spacy tool~\footnote{https://spacy.io/}. We calculate the frequencies of all verbs. We find that the long-tail distribution phenomenon of verbs is serious. Therefore, for each verb, we only keep at most $K$ events of it as the seed events. We denote the final seed event set as $\sS^{\gE}=\{(\gE_i, \gI_i)\}$.

% After the event diversification, the distribution of types and scenarios of events is largely enriched. This process improves our model's generalization of various events and enhances the scenario knowledge.

% % ===

\subsection{Event Diversification}
\label{ed}
Event diversification aims to curate a varied collection of seed events, encompassing multiple types and scenarios for ensuing evolutionary learning. We initiate this process with a corpus of text and image pairs $\{(\gE_i, \gI_i)\}$, where each pair jointly represents an event. We next extract the triggers to represent the events. Trigger words are typically verbs that explicitly signify the event’s occurrence~\cite{ace2004}. We employ the Spacy tool\footnote{https://spacy.io/} to identify the primary verb $\gV_{\gE_i}$ within each text $\gE_i$ as the trigger.

Observing a long-tail distribution in trigger frequency, we only include $K$ events per trigger to establish a balanced seed event set, denoted as $\sS^{\gE}=\{(\gE_i, \gI_i)\}$. The outcome of this event diversification step is more diversified event types and scenarios, thereby broadening our model’s generalization capabilities and strengthening its understanding of varied contexts.

\begin{figure}[!tb]
    \centering
    \includegraphics[width=1\columnwidth]{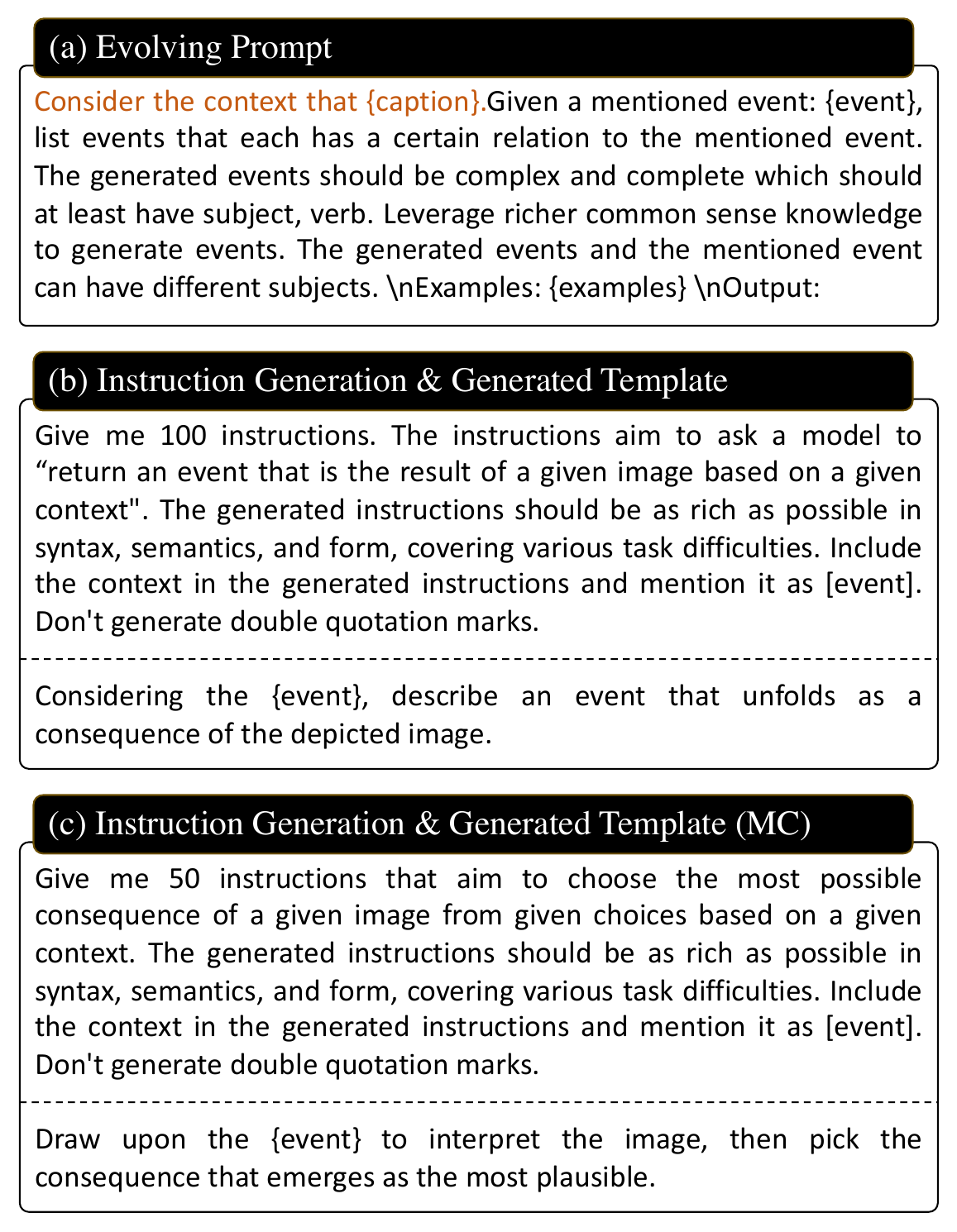}
    \caption{(a) Evolving prompt. The sentence in brown only exists if $\gE$ is the seed event. In such a case, we add the caption of $\gI$. (b) Instruction templates generation of \texttt{Result} relation and one example of generated template. (c) Multiple-choice Instruction templates generation of \texttt{Result} relation and one example of generated template. \{caption\} is the placeholder for the image caption. \{event\} and \{examples\} are for the event $\gE$ and in-context examples. }
    \label{fig:prompts}
\end{figure}

\begin{algorithm}[!t]
\small
\caption{Event Graph Evolution algorithm.}
\label{egea}
\SetKwInOut{KwIn}{Input}
\SetKwInOut{KwOut}{Output}
\KwIn{Seed event $\gE$ and the caption $\gC$, evolving relations $\sR^{E}$, evolving steps $L$.}
\KwOut{Event-evolving graph $\sG$.}
\SetKwFunction{AddNode}{AddNode}
\SetKwFunction{AddEdge}{AddEdge}
\SetKwFunction{SampleRel}{SampleRel}
\SetKwFunction{Evolve}{Evolve}
\SetKwFunction{SampleEvent}{SampleEvent}
\SetKwFunction{Append}{Append}

$\sG.\AddNode(\gE)$, $\Tilde{\mathbb{E}} = [\gE]$

\For{$i \leftarrow 1$ \KwTo $L$}{
    $\sN = [\ ]$ \\
    \For{$\gE_{j}$ \textbf{in} $\Tilde{\mathbb{E}}$}{
        \eIf{$i == 1$}{
            $\{(\gE_k, \gR_k)\} = \Evolve(\gE_{j}, \gC, \SampleRel(\sR^{E}, 2))$ 
        }
        {
            $\{(\gE_k, \gR_k)\} = \Evolve(\gE_{j}, \SampleRel(\sR^{E}, 2))$ 
        }

        \For{$\gE_{k}, \gR_k$ \textbf{in} $\SampleEvent(\{(\gE_k, \gR_k)\}, 2)$}{
        $\sG.\AddNode(\gE_{k})$ \\
        $\sG.\AddEdge(\gE_{j}, \gR_k, \gE_{k})$ \\
        $\sN.\Append(\gE_k)$\\
        }
    }
    $\Tilde{\mathbb{E}} = \sN$ \\

}
\KwRet{$\sG$}

\end{algorithm}

\subsection{Event Graph Evolution}
\label{ege}
For the goal of enhancing the comprehension of event evolution, we utilize the seed events $\sS^{\gE}$ to construct event-evolving graphs through our designed event graph evolution methodology. Building on insights from prior work where LLMs like ChatGPT\footnote{https://openai.com/} have demonstrated proficiency in generating coherent event narratives~\cite{gunjal2023drafting, li2023opendomain}, we apply a breadth-first search (BFS) strategy using the ChatGPT to expand each seed event $(\gE, \gI) \in \sS^{\gE}$ both forward and backward in event happening time.  We show the process of either direction in Algorithm~\ref{egea}.

We introduce the process for the forward evolution. Starting from the seed event $\gE$, we consider forward-oriented relations such as $\sR^{E} = \{\texttt{Result}, \texttt{After}, \texttt{HasIntention}\}$\footnote{Relations are directed from the generated to the queried event, for instance, generating the \texttt{Result} for a given event. \texttt{HasIntention} implies the head event is intended by subjects in the tail event.}. For each iteration of this process, we invoke the ChatGPT to produce events consistent with sampled relations from $\sR^{E}$, as described in Equation~\ref{gen}. 
At the beginning, recognizing the potential bias of relying solely on textual events, we incorporate visual information of the seed event. Specifically, while evolving a seed event, we add its image caption to provide contextual details, promoting more accurate evolution. When evolving the intermediate events, we only use just their text. The prompt template for this evolution process is depicted in Figure~\ref{fig:prompts}(a).
After $L$ iterations, we acquire an event-evolving graph $\gG$.

Besides, we also consider backward evolution. Our motivation for that is intuitive. We want the model to cognize event evolution in an complete timeline including both directions. Since we always start from an intermediate event in the timeline, we need to perform both forward and backward evolution. 
To do that, we consider evolving relations $\sR^{E} = \{\texttt{Cause}, \texttt{Before}, \texttt{IsIntention}\}$ and remains the other steps the same.

After the both sides evolution, we denote the outputs as the event-evolving graph $\gG$ which entails the rich evolution mechanism of the event scenario.

\subsection{Instruction Encapsulation}
\label{ie}
To endow the knowledge of the evolving graphs $\gG$ for model training, we turn to multi-modal instruction-tuning, a technique with proven efficiency in adapting models to human-like comprehension~\cite{zhu2023minigpt, sun2023generative, li2023mimic, liu2023visual,li2023otter,dai2023instructblip}. Our approach involves transforming the components of $\gG$, represented as $\gG=(\sV, \sW)$ with nodes $\sV$ and edges $\sW$, into instruction-tuning data.

For each node $\gE_i \in \sV$, we aim to create a datum comprising the seed event $\gE^{s}$, its associated image $\gI$, the relation $\gR_i$, and the event $\gE_i$. However, directly inferring $\gR_i$ between nodes $\gE^{s}$ and $\gE_i$ is not straightforward if they are non-adjacent. We address this by introducing induction rules that leverage the properties of inter-event relations, as detailed in Table~\ref{rule}. For example, in an evolving graph $\gG$, there exists a path from the seed event $\gE^{s}$ and another event $\gE_2$: $\gE^{s}$$\Rightarrow$[After]$\Rightarrow$$\gE_1$$\Rightarrow$[Result]$\Rightarrow$$\gE_2$.
According to rule 1 in Table~\ref{rule}: (After)$\star$(Result)+(After)$\star$ infers Result, where $\star$ denotes there exists zero or more, + means there is at least one. We induce $\gE^{s}$$\Rightarrow$[Result]$\Rightarrow$$\gE_2$.
By applying these rules, we derive the indirect relation $\gR_i$.

Then we embed all the data with our instruction-tuning templates to form an instruction tuning dataset. For the templates, to avoid the laborious task of manual template creation, we employ ChatGPT to generate diverse question templates for each relation type. With 100 templates from ChatGPT, the templates aim to reason about the tail event based on the provided visual and/or textual events in accordance with Equation~\ref{gen}. Considering the possible absence of textual input, we generate two variations for each of the $|\sS^{\gR}|$ relations: one with textual input and one without.

For any given data $(\gE^{s}, \gI, \gR_i, \gE_i)$, we randomly determine whether to include textual event information. We then match a suitable template to the relation type $\gR_i$ and encapsulate all the items into our instruction-tuning dataset. An example of an encapsulated datum is illustrated in Figure~\ref{fig:prompts}(b).

\begin{table}[!t]
\centering
\footnotesize

\setlength{\tabcolsep}{3mm}
\begin{tabular}{ll}

\toprule
\textsc{Rule}&\textsc{Induction}\\
\midrule
(\texttt{After})$\star$(\texttt{Result})+(\texttt{After})$\star$&\texttt{Result}\\
(\texttt{After})$\star$(\texttt{HasIntention})+( \texttt{After})$\star$&\texttt{HasIntention}\\
(\texttt{After})+&\texttt{After}\\
(\texttt{Before})$\star$(\texttt{Cause})+(\texttt{Before})&\texttt{Cause}\\
(\texttt{Before})$\star$(\texttt{IsIntention})+(\texttt{Before})&\texttt{IsIntention} \\
(\texttt{Before})+&\texttt{Before}\\

\bottomrule

\end{tabular}
\caption{Relation induction rules. $\star$ denotes there exists zero or more. + means there is at least one.}
\label{rule}
\end{table}

\subsection{Guiding Discrimination}
\label{eld}
To ensure accuracy during event graph evolution and guide the model away from generating erroneous events, we introduce a guiding discrimination training paradigm. This mechanism is pivotal in preventing the evolution process from producing hallucinations which is similar to DPO~\cite{rafailov2023direct}. In this paradigm, we task the model with identifying the correct event amongst a set of carefully selected negative events.

\begin{equation}
\label{mc}
    \gE^{t} = \mathrm{M}~(\gE, \gI, \gR, \sD), \quad \gR\in\sS^{\gR},
\end{equation}
where $\sD$ is the candidate set consisting of the correct event $\gE^{t}$ and a few negative events.

The discrimination training is challenging to perform due to the sourcing of these negative events. For which we formulate two negative event acquisition strategies:

\textbf{Semantic:} This strategy requires model to discriminate the semantic similar events. To forge semantically similar negative events, we first compile a pool of all events of the generated graphs. For any positive event $\gE$, utilizing Spacy for dependency parsing, we compute the tree edit distance and the word overlap rate between $\gE$ and each event in this pool\footnote{https://github.com/timtadh/zhang-shasha}. Filtering by the preset thresholds for these metrics, we select the top two events that are close to $\gE$. This method sharpens the model's ability to distinguish between events with closely related linguistic structures.

\textbf{Evolving:} This strategy enhances the model's grasp on the directionality of event evolution. Leveraging the bidirectional nature of our event generation, namely forward and backward directions, we select two negative events from the opposite direction of the positive event's evolution. These negatives are particularly challenging as they maintain shared arguments within the same scenario but differ in their logical sequence. This practice further refines the model's reasoning skills for establishing the correct evolution path.

From the total four negative events generated through these strategies, we randomly select two of them. These, alongside the correct event, are then encapsulated into a multiple-choice format. We also create diverse multiple-choice question templates for each relation type via ChatGPT. An example of such a generation prompt and a corresponding template is presented in Figure~\ref{fig:prompts}(c). Comprehensive statistics of the dataset are detailed in Table~\ref{stats}.

\begin{table}
% \vspace{-4mm}
% \setlength{\belowcaptionskip}{-6mm}
% \setlength{\abovecaptionskip}{-0.1mm}
\centering
\small
\setlength{\tabcolsep}{2.5mm}{\begin{tabular}{lccc}

\toprule
  \textsc{Graph} & \textsc{Node} & \textsc{Trainset} & \textsc{Avg Input Token}\\
\midrule
\midrule

3600 & 38.36& 7470& 104.17\\

 \bottomrule
\end{tabular}}
\caption{Trainset statistics.}
\label{stats} 
\end{table}
% \vspace{-5mm}

\subsection{Training}
\label{train}
After acquire both MMER and guiding discrimination dataset, we finetune the backbone by combining the MMER loss $\gL^{R}$~(from Eq.\ref{gen}) and the guiding discrimination loss $\gL^D$~(from Eq.\ref{mc}):

\begin{equation}
\setlength\abovedisplayskip{1pt}
\setlength\belowdisplayskip{1pt}
\begin{aligned}
    \gL^{R} &= -\sum_{(\gE, \gI, \gR)} \log\displaystyle P(\gE^{t}|\gE, \gI, \gR), \\
    \gL^D &= -\sum_{(\gE, \gI, \gR, \sD)} \log\displaystyle P(\gE^{t}|\gE, \gI, \gR, \sD), \\
    \gL\ &= \gL^{R} + \gL^{D}
\end{aligned}
\end{equation}

\section{Experiments}

% \vspace{-1mm}

\subsection{Construction of \mev}
% We incorporate nine relational reasoning test datasets to testify our method covering event-related visual question answering, visual event prediction, and storytelling. 
% In this work, we denote the multiple-choice tasks as \textsc{Close} tasks and tasks without candidates as \textsc{Open}.
% We include \texttt{VCOPA}~\cite{yeo2018visual}, \texttt{VisCa}~\cite{zhang2021learning}, \texttt{VisualComet}~\cite{park2020visualcomet}, \texttt{IgSEG}~\cite{huang2021igseg}, \texttt{VIST}~\cite{huang2016visual}. Detailed descriptions are in Appendix A. 

To comprehensively evaluate the models' abilities of MMER on diversified inter-event relations, we collect and curate a benchmark \mev.
It incorporates nine test sets covering event-related visual question answering (\texttt{VCOPA}, \texttt{VisCa}, \texttt{VisualComet}), visual event prediction (\texttt{IgSEG}), and storytelling (\texttt{VIST}). \mev~evaluates event relations of causality, temporality, and intent. Besides, \mev~also covers two reasoning paradigms that are multiple-choice close reasoning tasks (\textsc{Close}) and open reasoning without candidates (\textsc{Open}). We elaborate on the curation process as follows.

\noindent\texttt{VCOPA} This is the task of commonsense VQA~\cite{yeo2018visual}. Given an image $\gI$ and two candidate options, the task is to select a more plausible cause or effect option. We also transform this dataset into an open reasoning task in which we don't provide the candidates and require the model to generate the answer. We denote the original multiple-choice task as \texttt{VCOPA-C} and the transformed task as \texttt{VCOPA-O}.

\noindent\texttt{VisCa} This is a dataset of learning contextual causality from the visual and textual signals~\cite{zhang2021learning}. The original task is formulated as that given two images as the context and two textual sentence descriptions, models need to determine if the former sentence causes the latter one. We transform it into our VQA task. We keep the image and first sentence and regard the second sentence as the label to generate. We retrieve one negative sentence by the ground truth and form it as a multiple-choice task. We also adapt the multiple-choice task into an open reasoning similar to \texttt{VCOPA-O}. We denote these two tasks as \texttt{VisCa-C} and \texttt{VisCa-O}. 

\noindent\texttt{VisualComet} This is an open commonsense VQA task which is to answer situations before or after~\cite{park2020visualcomet}. We also retrieve a negative answer to formulate it into a multiple-choice task. We denote these two tasks as \texttt{VC-O} and \texttt{VC-C}. 

\noindent\texttt{IgSEG} This dataset aims to predict future events based on what has happened~\cite{huang2021igseg}. Specifically, given a sequence of sentences in sequential order and the image of what will happen next, the models need to generate a sentence for this image. In addition, we also retrieve one negative event and form it as a multiple-choice task. We denote these two tasks as \texttt{IgSEG-O} and \texttt{IgSEG-C}.

% \begin{table}
% % \setlength{\belowcaptionskip}{-6mm}
% % \setlength{\abovecaptionskip}{-0.1mm}

% \centering
% \footnotesize
% \setlength{\tabcolsep}{0.5mm}{\begin{tabular}{ccccccccc}

% \toprule
%   \textsc{VCOPA-C/O} & \textsc{VisCa-C/O} & \textsc{VC-C/O} & \textsc{IgSEG-C/O} & \textsc{VIST}\\
% \midrule
% \midrule

% 5.75/3.63 & 3.77/2.48 & 3.85/3.85 & 3.57/4.36 & 5.78 \\

%  \bottomrule
% \end{tabular}}
% \caption{Trainset leakage percentages.}
% \label{leakage} 

% \end{table}

\noindent\texttt{VIST} It's the storytelling task which is to generate the next story given the previous story in sentences and an image~\cite{huang2016visual}.

% We also compute the percentage of data leakage of our trainset to the test set. We regard test data as a leakage if the train set has at least one data that has more than 0.5 cosine similarity to it. We report the percentage in Table~\ref{leakage}. There's only a small portion of data leakage. 

\begin{table*}[!t]
\centering
\small
\setlength{\tabcolsep}{1.2mm}
\begin{tabular}{lcccccc}
\toprule
$\clubsuit$ & \texttt{VCOPA-C} & \texttt{VisCa-O} & \texttt{VC-C} & \texttt{VCOPA-O} & \texttt{VisCa-O} & \texttt{VC-O}  \\
\cmidrule(lr){2-7}
& \multicolumn{6}{c}{\textsc{VQA}} \\

\midrule
\midrule
InstructBLIP~\cite{dai2023instructblip} & 63.33 & 64.78 & 51.25 & 7.57~/~2.31~/~9.32 & 7.56~/~1.01~/~14.87 & 12.30~/~\textbf{4.84}~/~13.72 \\
Otter~\cite{li2023otter} & 57.27 & 55.97 & 45.10 & 11.78~/~1.35~/~17.12 & 10.29~/~0.51~/~10.51 & 7.96~/~3.18~/~9.13 \\
LLaVA-Lora~\cite{liu2023visual} & 46.06 & 45.28 & 
45.60 & 7.66~/~1.44~/~0.64 & 7.06~/~0.67~/~5.66 & 7.57~/~2.31~/~3.32\\
MiniGPT-4~\cite{zhu2023minigpt} & 56.67 & 47.80 & 51.40 & 9.78~/~2.44~/~7.05 & 7.87~/~1.55~/~10.30 & 6.92~/~1.78~/~0.42  \\
MiniGPT-4-v2~\cite{chen2023minigpt} & 49.70 & 52.83 & 54.60 & 8.90~/~2.13~/~2.09 & 8.89~/~1.21~/~8.55 &  7.54~/~3.03~/~5.06\\

\me~(Ours) & \textbf{66.06} & \textbf{72.33} & \textbf{68.10} & \textbf{19.18}~/~\textbf{2.92}~/~\textbf{26.02} & \textbf{19.16}~/~\textbf{3.40}~/~\textbf{29.58} & \textbf{16.28}~/~3.99~/~\textbf{22.93} \\
\bottomrule
\end{tabular}
\caption{Main results of VQA tasks. The bold number represents the highest score.}
\label{vqa}
\end{table*}

\begin{table}[!t]
\centering
\footnotesize
\setlength{\tabcolsep}{0.5mm}
\begin{tabular}{lccc}
\toprule
$\clubsuit$ & \texttt{IgSEG-C} & \texttt{IgSEG-O} & \texttt{VIST} \\
\cmidrule(lr){2-4}
& \multicolumn{2}{c}{\textsc{Prediction}} & \textsc{StoryTelling} \\
\midrule
\midrule
InstructBLIP & 55.10 & 8.13/\textbf{2.63}/15.91 & 6.71/1.22/11.31 \\
Otter & 53.20 & 7.57/1.35/4.34 & 7.63/1.20/10.51  \\
LLaVA-Lora & 46.40 & 9.03/1.50/4.46 & 9.09/\textbf{3.03}/5.53  \\
MiniGPT-4 & 49.90 & 8.72/1.54/3.24 & 8.66/1.67/9.64  \\
MiniGPT-4-v2 & 51.30 & 8.69/1.45/3.73 & 8.95/1.68/10.44  \\

\me~(Ours) & \textbf{66.50} & \textbf{14.00}/1.41/\textbf{19.41} & \textbf{14.38}/1.44/\textbf{25.60}  \\
\bottomrule
\end{tabular}
\caption{Main results of visual event prediction and storytelling. The bold numbers represent the best score.}
\label{ps}
\end{table}

\subsection{Baselines}
We compare baselines as LLaVA-Lora~\cite{hu2021lora}, InstructBLIP~\cite{dai2023instructblip}, Otter~\cite{awadalla2023openflamingo}, MiniGPT-4~\cite{zhu2023minigpt}, MiniGPT-4-v2~\cite{chen2023minigpt}. We show more details in Appendix~\ref{baselines}.

% \vspace{-1mm}

\subsection{Implementation Settings}
We use InstructBLIP~\cite{dai2023instructblip} to generate the image captions for event graph evolution. We sample two evolving events for each event in the BFS. We set the evolution steps as 3. We finally constructed 15,000 instruction-tuning data. 

For our model, we use LLaVA-v1.3 after the first pre-training stage as our backbone~\cite{liu2023visual} and train with Lora~\cite{hu2021lora}, making it comparable to the LLaVA-Lora-v1.3-7B baseline. We use deepspeed\footnote{https://www.deepspeed.ai/}, zero-2 without CPU offloading. We set the batch size to 16 on 4$\times$V100 GPUs.

In pilot experiments, we conducted tests with multiple input prompts for each model in order to identify the most effective prompts for evaluation. Despite variations in prompts, we observed only minimal fluctuations in the results. To ensure consistency and mitigate the other influences, we maintained uniformity by using the same prompt for all models performing a task. Detailed prompts can be located in the Appendix~\ref{inference prompts}. For the multiple-choice tasks, we transformed them into multiple-choice questions and instructed the model to respond with the corresponding label of choice. For \textsc{Close} tasks, we design an answer decoding strategy and show in Algorithm~\ref{decode}. We find this strategy can handle almost all situations.

\begin{algorithm}[t]
\small
\SetAlgoNoEnd
\SetKwInOut{Input}{Input}
\SetKwInOut{Output}{Output}
\SetKw{KwTo}{in}
\SetKwIF{If}{ElseIf}{Else}{if}{then}{else if}{else}{endif}

\Input{Prediction $\gP$, candidate set $\sD$.}
\Output{Answer $\gA$.}

pattern = \text{"the(?: correct)? (?:option$|$answer) is[}\textbackslash\text{ s:]+([A-H])"} \\
\If{$\gP$$\mathrm{.startsWithAlphabet()}$}{
    $\gA$ = starts\_alphabet
}
\uElseIf{
$\mathrm{re.match(pattern}$, $\gP$)
}
{
    $\gA$ = $\mathrm{re.extract(\gP, patten)}$
    % Extract the \textit{Answer} follow the \textit{pattern} from \textit{Prediction}.
}
\Else{
    $\gA$=$\underset{c \in \sD}{\mathrm{argmax}}$($\mathrm{WordOverlap}$($c$, $\gP$)
}
\KwRet{$\gA$}
\caption{\textsc{Close} answer decoding.}
\label{decode}

\end{algorithm}

% \vspace{-1mm}

\subsection{Evaluation Metrics}
For multiple-choice tasks, we employ accuracy as the metric. For \textsc{Open} tasks we utilize BLEU-1/2~\cite{papineni2002bleu} and BERT-SCORE~\cite{zhang2019bertscore} as measures. 

\begin{figure*}

    \begin{subfigure}{0.5\textwidth}  % 创建第一个子图，占页面宽度的一半
        \centering
        \includegraphics[width=\linewidth]{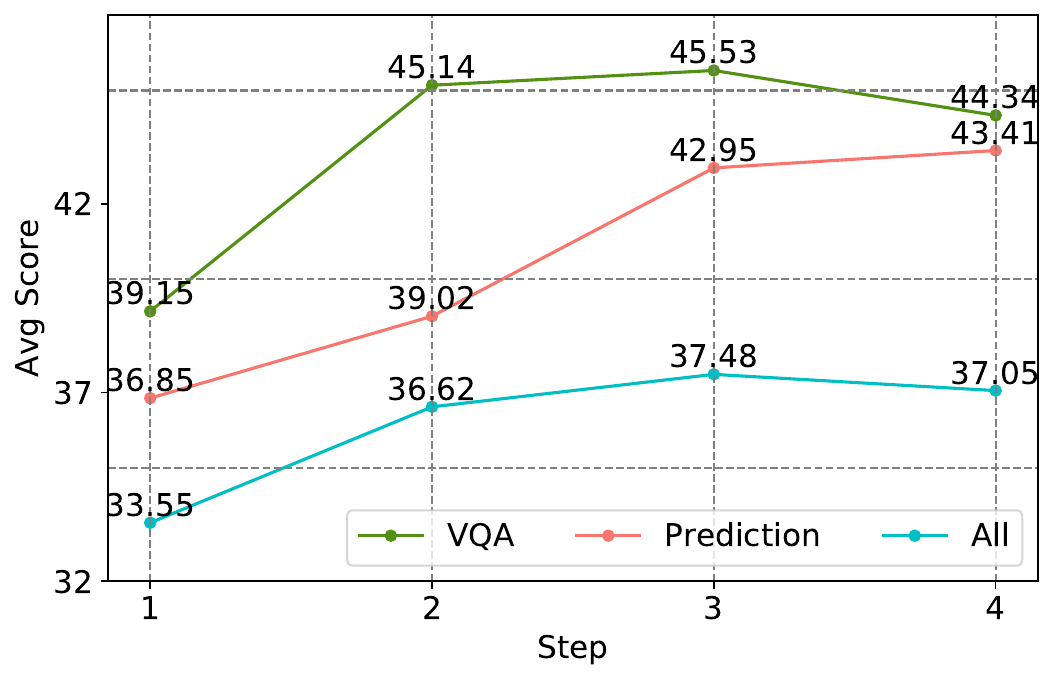}  % 插入第一个图像
        \caption{Average scores on \textsc{VQA}, \textsc{Prediction}, and the average of all results. }  % 添加子图标题
        \label{fig:subfig1}  % 添加子图标签，用于引用
    \end{subfigure}
    \begin{subfigure}{0.5\textwidth} % 创建第二个子图，占页面宽度的一半
        \centering
        \includegraphics[width=\linewidth]{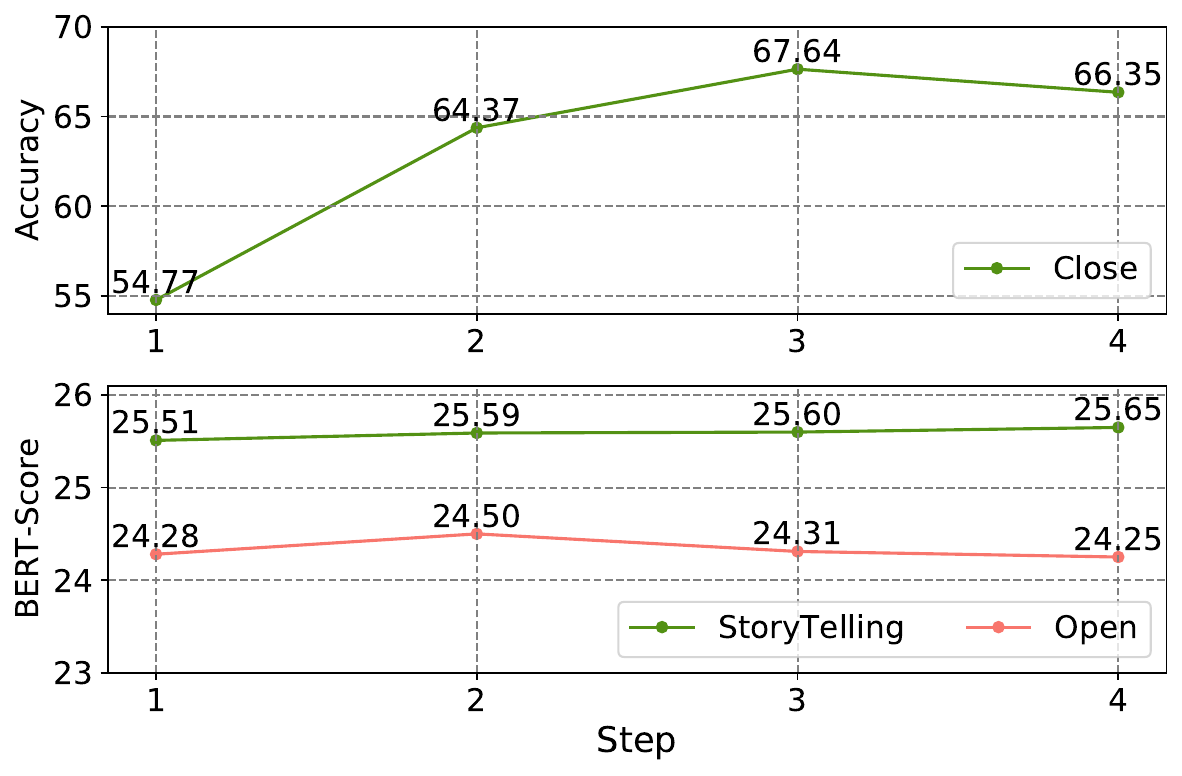}  % 插入第二个图像
        \caption{Average scores on \textsc{StoryTelling}, all \textsc{Close} tasks, and all \textsc{Open} tasks. }  % 添加子图标题
        \label{fig:subfig2}  % 添加子图标签，用于引用
    \end{subfigure}
    \caption{Analysis of steps of event graph evolution. }
    \label{fig:hop}  % 添加总图标签，用于引用
\end{figure*}

\subsection{Main Results}
We test our model on \mev~benchmark. We show the VQA results in Table~\ref{vqa}, visual event prediction and visual storytelling in Table~\ref{ps}. We calculate the various kinds of average scores in Table~\ref{avg}.

\noindent\textbf{\me~can effectively enhance performances of \textsc{VQA}.} \me~achieves the highest scores on three \textsc{Close} VQA, namely \texttt{VCOPA-C}, \texttt{VisCa-C}, and \texttt{VC-C} in Tabel~\ref{vqa}. The results indicate \me~can distinguish the right events since the improvements from event graph evolution with guiding discrimination. 
For the three \textsc{Open} VQA datasets, among all metrics, BERT-SCORE can mostly evaluate the answering quality. We find \me~outperforms all other baselines to a large extent. These results demonstrate the effectiveness of our method on \textsc{Open} VQA. We also notice the BLEU-1/2 of \me~is higher than almost all models. Since BLEU-1/2 measures lexical similarity, \me~can generate more well-formed events as the ground truth. 
In all, our method improves the MMER.

\begin{table}[!t]
\setlength{\belowcaptionskip}{2mm}
\centering
\footnotesize
\setlength{\tabcolsep}{0.6mm}{
\begin{tabular}{lcccccc}

\toprule
$\clubsuit$  & \textsc{VQA} & \textsc{Pred} & \textsc{Story} & \textsc{Open} & \textsc{Close} & \textsc{All}  \\
\midrule
\midrule

InstructBLIP &33.01&35.50&11.31&12.53&54.11&25.16 \\
Otter &28.40 & 28.77 & 10.51&9.66&49.06&21.64 \\ 
LLaVA-Lora &23.92&25.43&5.53&4.64&45.85&17.17 \\ 
MiniGPT-4-v2 &26.49&26.57&9.64 & 6.44&51.30&20.08 \\
MiniGPT-4 & 28.86&27.51&10.44&7.84&53.11&21.60 \\
\me~(Ours) & \textbf{45.53}&\textbf{37.95}&\textbf{25.60}&\textbf{23.06}&\textbf{67.64}&\textbf{36.61} \\

 \bottomrule
\end{tabular}}
\caption{Various kinds of average results. The bold numbers represent the best score. \textsc{Pred} stands for visual event prediction. \textsc{Story} is visual story telling. \textsc{Close} and \textsc{Open} are close and open reasoning tasks respectively. \textsc{All} is the average performance on all test set. }
\label{avg} 
\end{table}

\noindent\textbf{\me~outperforms baselines in visual event prediction.}
\me~performs the best among all baselines in Table~\ref{ps}. The results demonstrate our training method enables the model to capture correct temporal relations leading to more precise prediction for the future. Compared to \textsc{VQA} tasks, We find all models perform worse in visual event prediction, indicating it needs more knowledge and reasoning ability to complete this task. 
In \textsc{Open} visual prediction, \me~also achieves the highest scores in BERT-SCORE. This shows our model can forecast semantic similar events. However, we find \me~performs slightly lower in BLEU-2 on \texttt{IgSEG-O}. Since BLEU calculates the 2-gram lexical similarity, this may indicate \me~can predict more diversified events with correct semantics rather than words merely in the context.

\noindent\textbf{\me~can generate advanced story.}
In Table~\ref{ps}, we find \me~can excel all baselines in \textsc{VIST}. The results show \me~can tell better stories by capturing more scenario knowledge and comprehending the inter-event relations. The event graph evolution affects the training of the model to acknowledge enriched event information rather than merely shallow step reasoning.

\noindent\textbf{In all, \me~can significantly improve the performance of the downstream tasks attributed to boosted capabilities of MMER.}
In Table~\ref{avg}, \me~excels all baselines on the average score of all datasets demonstrating the effectiveness of our method. Our event graph evolution process stimulates the contextual understanding of events. The guiding discrimination further mitigates the hallucinations of event reasoning yielding better performances. 

Among all relation types, the improvements of \textsc{VQA} and \textsc{StoryTelling} are larger than \textsc{Prediction}. It indicates our method benefits more for these tasks. \textsc{Prediction} is the hardest to learn attributed to its demand for pertaining for more abundant knowledge of events.

\begin{figure*}[!tb]
    \centering
    \includegraphics[width=2\columnwidth]{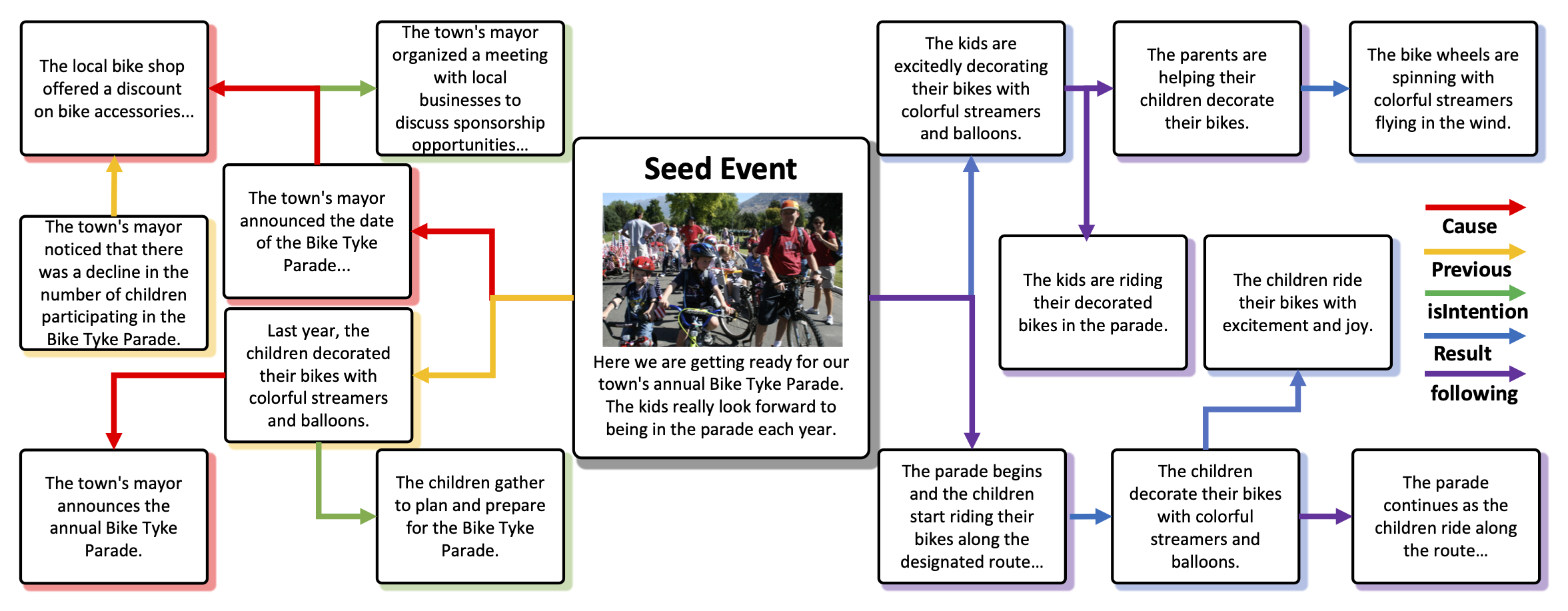}
    \caption{An example of an event-evolving graph. The event pointed to by the head cut is a tail event generated that satisfies the color relationship of the head cut.}
    \label{fig:case}
\end{figure*}

% \vspace{-2mm}

\begin{table}[!t]
\setlength{\belowcaptionskip}{2mm}
\centering
\footnotesize
\setlength{\tabcolsep}{0.7mm}{\begin{tabular}{lccccc}

\toprule
$\clubsuit$  & \texttt{VCOPA-O} & \texttt{VisCa-O} & \texttt{VC-O} & \texttt{IGSEG-O} & \texttt{VIST} \\
\midrule
\midrule

\me~w.o. D & 19.63 & 21.78 & 21.79 & 18.83 & 24.67\\
\me& 26.02 & 29.58 & 22.93 &  19.41 &  25.60\\

 \bottomrule
\end{tabular}}
\caption{Ablation study. \me~w.o. D is our method without guiding discrimination.}
\label{ablation} 
\end{table}

\subsection{Analysis}

\noindent\textbf{Evolution steps.} We conduct experiments on different evolution steps to verify the effectiveness of event graph evolution. We tested steps 1-4 respectively and calculated various average scores. We show the results in Figure~\ref{fig:hop}. 

As the average of all results, the performance of \me~increases from steps 1 to 3 in Figure~\ref{fig:hop} (a). This is consistent with our motivation that the event graph evolution enables the model to learn the rich knowledge of event evolution. Then, the model can complete MMER better.

We find the performances drop when the step is too large, namely larger than four. This may be attributed to the semantic drift of the event graph evolution. ChatGPT would generate less relevant content compared to the seed event if it evolves further. We find that the drop is most obvious in \textsc{VQA}, which may be probably due to \textsc{VQA} being the most strict relation among all event interrelations.

We find \me~can achieve a high score for \textsc{StoryTelling} when the evolution step is only one in Table~\ref{fig:hop} (b). \me~is 25.51 BERT-Score while InstructBLIP is 11.31. As the number of steps increases, \me~maintains a high score. This indicates that \me~completes the \textsc{StoryTelling} even on few evolution steps.

\noindent\textbf{Effect of guiding discrimination.} We ablate guiding discrimination and show the results in Table~\ref{ablation}. We find that all performances drop if \me~trains without guiding discrimination. It indicates that discrimination can guide the evolution and mitigate hallucinations.

\begin{figure}[!tb]
\setlength{\belowcaptionskip}{1mm}
    \centering
    \includegraphics[width=1\columnwidth]{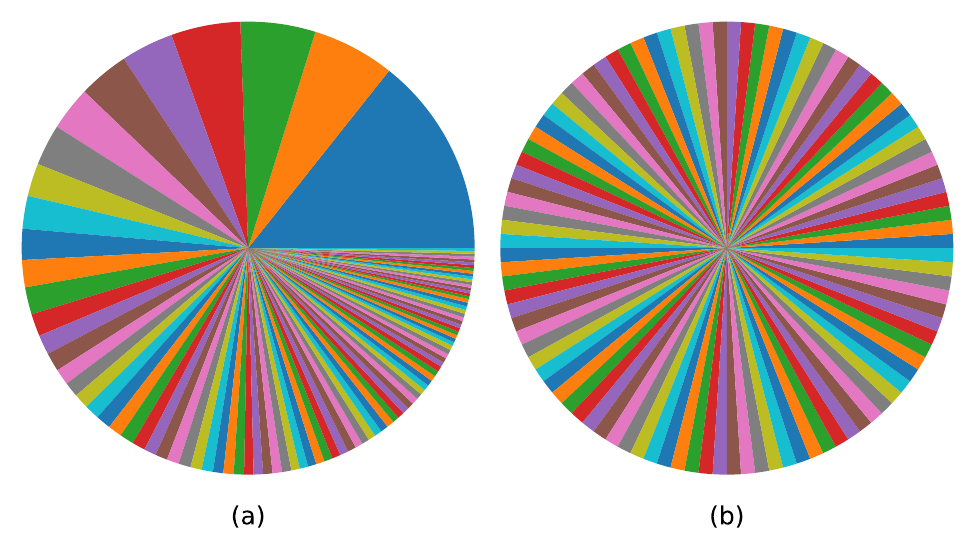}
    \caption{Distribution of verbs before and after event diversification. Each part of the pie chart is the proportion of a verb. We present the 100 most frequent verbs with and without event diversification. (a) w.o. event diversification. (b) w.t. event diversification.}
    \label{fig:dist}
\end{figure}

\noindent\textbf{Examples of event graph evolution.} We showcase two examples of event graph evolution in Figure~\ref{fig:case}. We find our evolving graphs can sufficiently contain information and knowledge of event scenarios. With the aid of event-evolving graphs, \me~learns more abundant event knowledge and relation inter-connections.

\noindent\textbf{Effect of event diversification.} We compute the event verb distribution. We show two verb distributions with or without event diversification. The results are in Figure~\ref{fig:dist}. We find the distribution is significantly diversified after the event diversification process. It enables \me~to be trained in various event scenarios and domains. 
% \vspace{-2mm}
\section{Relation Works}
% \vspace{-2mm}
\noindent\textbf{Multi-Modal Event Relational Reasoning}
As one of the relation types, causality reasoning is crucial for exploring the cause and effect of events~\cite{yeo2018visual,zhang2021learning, chadha2021ireason,ignat2021whyact}. 
% \citet{yeo2018visual} detect the reason or result of an image of an event. \citet{zhang2021learning, chadha2021ireason} proposes methods for training contextual-aware visual causality identification. \citet{ignat2021whyact} describe a multi-modal model that leverages visual and textual information to automatically infer the reasons corresponding to an action presented in the video. 
Apart from causality, event temporal reasoning forms a basic ability~\cite{zellers2019recognition, park2020visualcomet, zellers2021merlot}.  
% \citet{zellers2019recognition} model event temporal knowledge of grounded subjects in the images. \citet{park2020visualcomet} predict events happening before or after. \citet{zellers2021merlot} trains the model not only to match images to temporally corresponding words but also to contextualize events happening globally over time. 
Event intentional reasoning uncovers the intentions of the subjects of the events~\cite{park2020visualcomet, li2023intentqa}. 
% \citet{park2020visualcomet} infer the intention of a grounded person in an image. \citet{li2023intentqa} propose a context-aware video Intent reasoning model to align the correct intents. 
Besides, there exists research on other relation types as well~\cite{kim2022cosim, hessel2022abduction}. 
% \citet{kim2022cosim} model the event counterfactual reasoning which is to query what would happen if there was a changed situation. \citet{hessel2022abduction} reason abductively and hypothesize about what lies beyond the literal content of an image.
Multi-modal event relational reasoning constitutes a foundational capability for a range of downstream tasks in the realm of multi-modal reasoning. Our research endeavors to further enhance this crucial skill.

% \noindent\textbf{Applications of MM Event Relational Reasoning}
% MM Event Relational Reasoning contributes to MM applications. \citet{huang2016visual,hsu2020knowledge,xu2021imagine,hessel2022abduction,yu-etal-2017-hierarchically,smilevski2018stories} generate stories based on MM background information. The quality of story generation highly depends on event relational reasoning since it requires the implicit understanding of correlations of events in stories. \citet{ji2022vscript} introduce visual script learning that is to plot the script lines from MM signals. It demands temporal and causal reasoning the evolve the scripts. \citet{huang2021igseg} forecast future events based on what has already happened represented in images and words. \citet{liu2023video} model videos and construct event timelines under the scenario. MM event relational reasoning plays a crucial role in such tasks.

\noindent\textbf{Multi-Modal Instruction tuning} With the significant success of instruction tuning~\cite{ouyang2022training,xu2024survey}, current research has extended its capability to multi-modality. MM instruction tuning trains the model the follow instructions for questions about the images. Compared to textual instruction tuning, harvesting MM data with instructions is tougher. \citet{zhu2023minigpt} trains MiniGPT-4 by further aligning pretrained EVA-CLIP~\cite{fang2023eva} and Vicuna~\cite{vicuna2023}. \citet{liu2023visual} generate visual instruction data by requiring ChatGPT/GPT-4 with the given image and its caption. \citet{dai2023instructblip} adapt human-labeled dataset into instruction data with pre-made templates. \citet{li2023mimic} construct in-context learning data with instructions and use this dataset to train an MM LLM.
These methods merely model shallow event evolving situations leading to poor ability of MM event relational reasoning.

\noindent\textbf{Script Induction}
Script induction is to induce or generate chains or graphs of events representing the evolving mechanism. \citet{du2022resin} induces 11 scripts of newsworthy scenarios from documents. 
\citet{gunjal2023drafting} attempt to generate event chains by querying large language models.
\citet{zhang2023human} constructs scripts by designing interactions between humans and LLM.
\citet{li2023opendomain} create event graphs in a pipeline operation with generation, ordering, and verification.
In our work, we are the first to utilize the ability of script induction from ChatGPT to construct our MM event-oriented instruction-tuning data. We expect our work may shed light on other event-oriented approaches.
% \vspace{-2mm}

\section{Conclusion}
% \vspace{-2mm}

We propose the Multi-Modal Event Evolution Learning for MMER. We design the event graph evolution process based on the diversified seed events. We then encapsulate the evolving graphs into instruction-tuning data. We introduce the guiding discrimination training paradigm to further improve the learning of evolution. We conduct experiments on our collected and curated \mev~benchmark for MMER. Results show the effectiveness of \me~and it achieves competitive performance among open-source visual instruction-tuning baselines.

\section*{Limitations}
Our method is limited to MMER of a single image. However, a more complex MMER may contain several images to express a scenario. We leave the construction of methods and benchmarks of this complex MMER to future work.

\bibliography{main}
\appendix
\clearpage
\section*{Appendix}

\section{Baselines}
\label{baselines}

\noindent\textbf{LLaVA-Lora} This is a MLLM trained on visual instruction-tuning. It's based on the visual encoder ViT-L/14-336px~\cite{radford2021learning} and the textual chat LLM vicuna-v1.3-7b~\cite{vicuna2023}. In the first pre-train stage, it is trained with image-text pairs. In the second stage, it is finetuned by LLM-generated instruction-tuning data with LoRA~\cite{hu2021lora}.

\noindent\textbf{InstructBLIP} It uses BLIP-2~\cite{li2023blip} framework as its foundation, InstructBLIP strategically restructures 26 pre-trained public datasets, including image captioning and VQA, into a format conducive to instruction tuning~\cite{dai2023instructblip}.

\noindent\textbf{Otter} This model combines multi-modal in-context learning with multi-modal instruction tuning, building upon the foundation of OpenFlamingo~\cite{awadalla2023openflamingo}. This involves updating the perceiver module and relevant components of the LLM throughout the training process. The instructional data is sourced from reputable datasets including VQAv2~\cite{antol2015vqa}, GQA~\cite{hudson2019gqa}, LLaVA, as well as a proprietary video dataset not available to the public.

\noindent\textbf{MiniGPT-4} This model conducts visual instruction tuning on the pre-trained BLIP-2~\cite{li2023blip}, specifically focusing on updating the linear layer~\cite{zhu2023minigpt}. The instructions primarily draw from the domain of image captioning tasks.

\noindent\textbf{MiniGPT-4-v2} This model performs as a unified interface to complete various tasks such as VQA, visual grounding, and image caption~\cite{chen2023minigpt}. Different from MiniGPT-4, it adds task identifiers into the prompt to guide the task completion. The backbone of MiniGPT-4-v2 is LLama-2~\cite{touvron2023llama}.

% \clearpage

% \section{Decoding Protocol}
% \label{decoding protocol}
% We show our decoding protocol for extracting answers of \textsc{Close} tasks as in Algorithm~\ref{decode}.

\section{Inference Prompts}
\label{inference prompts}

\begin{figure*}[!tb]
    \centering
    \includegraphics[width=2\columnwidth]{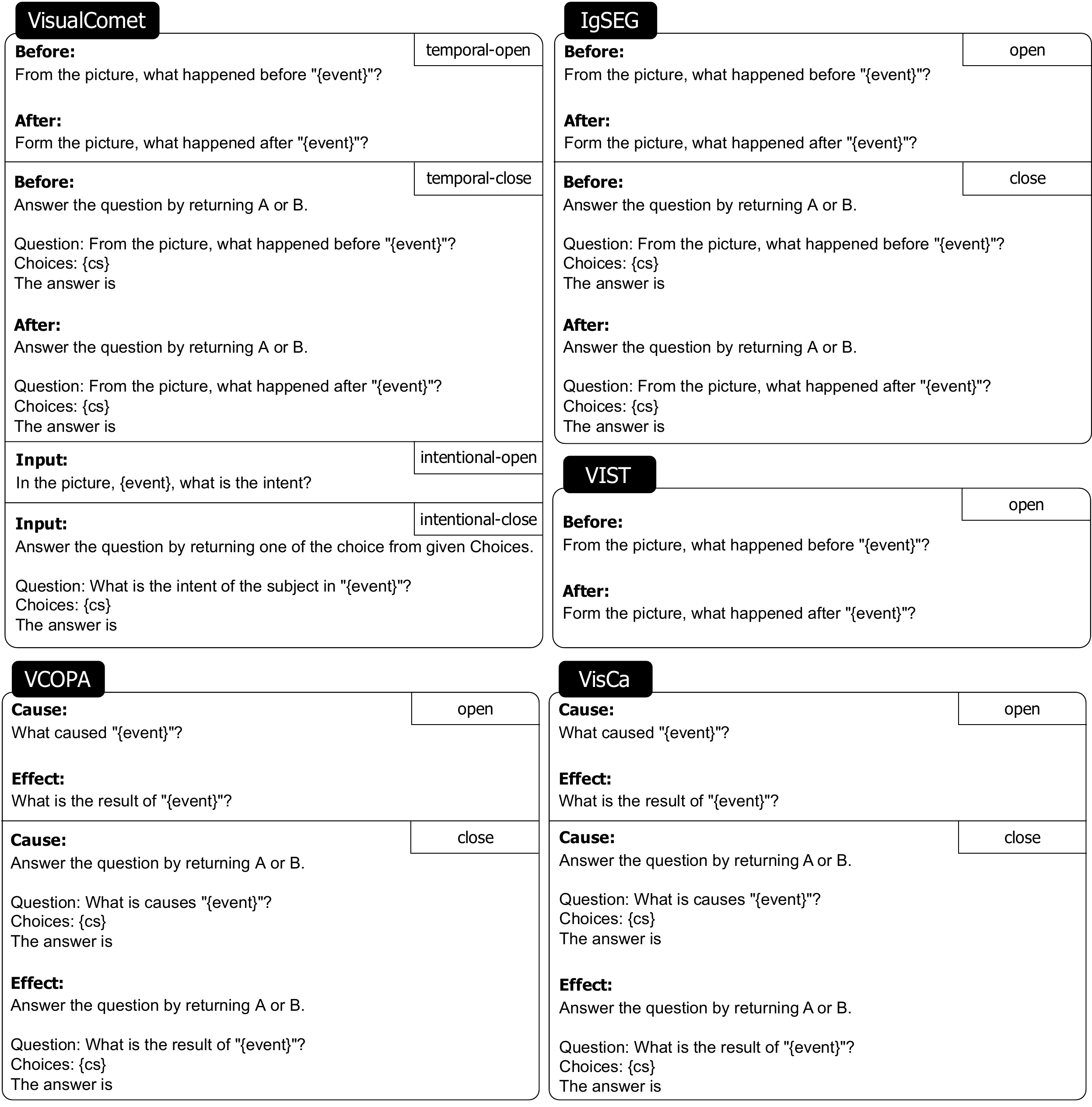}
    \caption{Inference prompts of all test set.}
    \label{fig:infer}
\end{figure*}

We show inference prompts of all test set in Figure~\ref{fig:infer}. We test various prompts in our pilot experiments and choose the prompts shown in Figure~\ref{fig:infer} which perform the best among others. We use the same prompts for all models.

\end{document}